\documentclass{article}
\usepackage[utf8]{inputenc}
\usepackage{hyperref} 
\usepackage{amsmath}
\usepackage{amsfonts}

\usepackage{booktabs}
\usepackage{pgfplots}
\usepackage{graphicx}
\usepackage{subcaption}
\usepackage{fancyhdr}
\usepackage[a4paper, margin=1in]{geometry}
\usepackage[most]{tcolorbox}
\tcbset{
  colback=gray!5,
  colframe=gray!40,
  top=2mm,
  bottom=2mm,
  left=2mm,
  right=2mm
}
\usepackage{listings}
\usepackage{xcolor}
\pgfplotsset{compat=1.18}

\title{\textbf{TCAndon-Router: Adaptive Reasoning Router for Multi-Agent Collaboration}}

\author{
\centering
\parbox{0.9\linewidth}{
\centering
\small
\textbf{Jiuzhou Zhao}$^{\ast}$,
\textbf{Chunrong Chen}$^{\ast}$,
\textbf{Chenqi Qiao}$^{\ast}$,
\textbf{Lebin Zheng}$^{\ast}$,\\
\textbf{Minqi Han}$^{\ast}$,
\textbf{Yanchi Liu}$^{\ast}$
\textbf{Yongzhou Xu}$^{\ast}$
\textbf{Xiaochuan Xu}$^{\ast}$
\textbf{Min Zhang}$^{\ast}$\\
\small
\medskip$^{\ast}$Tencent Cloud Andon \\
\medskip\{joskazhao, charentchen, chenqiqiao, lebinzheng\}@tencent.com \\
\{minqihan, yanncyliu, alanxu, xxcxu, alexzmzhang\}@tencent.com
}
}

\date{}

\begin{document}

\maketitle
\thispagestyle{fancy}

\begin{abstract}

\noindent Multi-Agent Systems(MAS) have become a powerful paradigm for building high performance intelligent applications. Within these systems, the router responsible for determining which expert agents should handle a given query plays a crucial role in overall performance. Existing routing strategies generally fall into two categories: performance routing, which balances latency and cost across models of different sizes, and task routing, which assigns queries to domain-specific experts to improve accuracy. In real-world enterprise applications, task routing is more suitable; however, most existing approaches rely on static single-label decisions, which introduce two major limitations: (i) difficulty in seamlessly integrating new agents as business domains expand, and (ii) routing conflicts caused by overlapping agent capabilities, ultimately degrading accuracy and robustness.To address these challenges, we propose TCAndon-Router(TCAR): an adaptive reasoning router for multi-agent collaboration. Unlike traditional routers, TCAR supports dynamic agent onboarding and first generates a natural-language reasoning chain before predicting a set of candidate agents capable of handling the query. In addition, we design a collaborative execution pipeline in which selected agents independently produce responses, which are then aggregated and refined into a single high-quality response by a dedicated Refining Agent.Experiments on public datasets and real enterprise data demonstrate that TCAR significantly improves routing accuracy, reduces routing conflicts, and remains robust in ambiguous scenarios. We have released TCAR at \url{https://huggingface.co/tencent/TCAndon-Router} to support future research on explainable and collaborative multi-agent routing.

\end{abstract}

\section{Introduction}

As large language models (LLMs) continue to improve in agent-based scenarios, multi-agent systems (MAS) have become increasingly mature. Their core idea is to decompose complex problems into smaller sub-tasks, each of which is handled by a specialized expert agent \cite{8352646}. Representative applications include Vibe Coding \cite{meske2025vibe} and Deep Research \cite{openai_deepresearch2025}.Today, large-scale enterprise systems—such as those in finance, healthcare, education, and cloud computing are also increasingly relying on multi-agent architectures to deliver accurate and efficient problem solving capabilities. Despite their broad applicability, MAS still face numerous challenges, including agent collaboration \cite{tran2025multi}, routing \cite{tran2025arch}, and security \cite{zhang2024agent}. As these ecosystems become more complex, routing becomes particularly crucial: it determines which agent should handle each specific task and therefore sets the lower bound for the overall performance of the MAS.

\begin{figure}
    \centering
    \includegraphics[width=0.9 \linewidth]{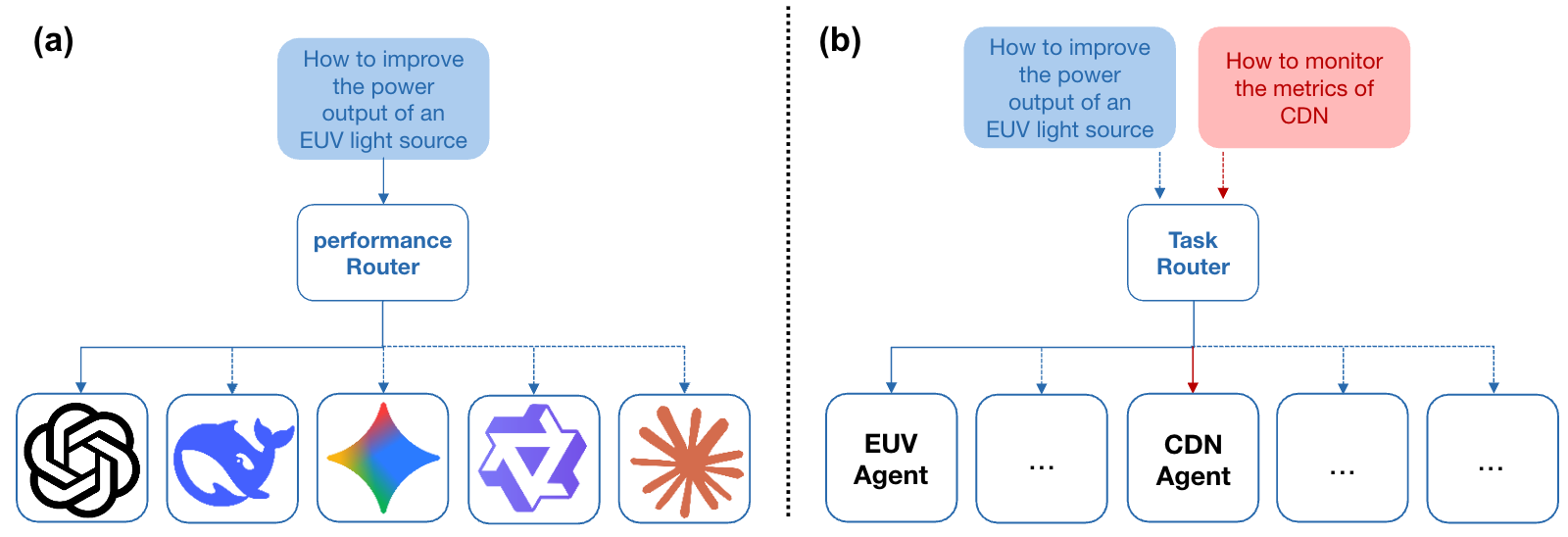}
    \caption{(a) performance router, which balances latency and cost across models of different sizes. (b) task router, which assigns queries to domain-specific experts to improve accuracy.}
    \label{fig:category}
\end{figure}

\medskip 
\noindent
Current routing strategies generally fall into two paradigms. The first is performance based routing, as depicted in ~\ref{fig:category}(a) which dynamically selects LLMs of different sizes based on the estimated difficulty of a query to balance precision and cost\cite{openai2025gpt5,jitkrittum2025universal,shnitzer2023large}. The second is task based routing, as shown in ~\ref{fig:category}(b) which assigns user queries to domain-specific expert agents to achieve higher task precision\cite{tran2025arch}. Performance based routing focuses on computational efficiency and model selection, whereas task based routing serves as the backbone of enterprise multi-agent systems by ensuring that queries are handled by domain experts rather than general-purpose models. Despite its strong performance, task based routing faces a critical limitation in real-world enterprise environments: \textbf{agent conflicts}. Unlike clean academic datasets, enterprise scenarios often involve overlapping agent responsibilities and multi-intent queries. For example, in cloud computing, an issue such as "website latency" can simultaneously be related to CDN configuration, public network quality, or application-layer bottlenecks. Traditional task based routers typically rely on single-label classification, which forces a single choice even when multiple experts are appropriate. This leads to routing errors, brittle behavior under ambiguous queries, and reduced system reliability. Furthermore, most routers support only static routing, making them difficult to adapt to dynamic category changes in enterprise environments. The lack of explicit reasoning during routing also limits interpretability and robustness.

\medskip 
\noindent
To address these challenges, we propose TCAR, adaptive reasoning router for multi-agent collaboration. Unlike existing routers that can only output a single label, TCAR generates structured natural language reasoning \cite{NEURIPS2022_9d560961} and identifies all expert agents that may be relevant to the current problem. This design transforms routing from a traditional black-box classification task into a transparent and interpretable reasoning process, enabling the system to articulate when and why multiple agents might be applicable. Beyond the agent selection mechanism, we further introduce a novel collaborative execution pipeline. Each selected expert agent first produces a domain-specific response, after which a Refining Agent \cite{jeong2025adaptive} aggregates and reconciles these outputs to form a logically coherent final answer. This architecture aligns with recent advances in collaborative multi-agent LLM systems, where domain experts provide complementary perspectives and a coordinator synthesizes the final decision \cite{grotschla2025agentsnet,talebirad2023multi}.This paradigm also mirrors human organizational workflows: multiple subject-matter experts contribute their insights, and a senior analyst ultimately integrates them into a unified conclusion.

\medskip 
\noindent
We evaluate TCAR on public benchmarks (CLINC150 \cite{larson2019evaluation}, HWU64 \cite{liu2021benchmarking}, MINDS14 \cite{DBLP:journals/corr/abs-2104-08524}, and SGD \cite{rastogi2020schema}) as well as on a private dataset from real-world cloud computing scenarios. Experimental results demonstrate that combining multi-agent routing with a refinement mechanism significantly improves accuracy, reduces routing conflicts, and enhances system robustness when handling ambiguous or cross-domain queries. 

\medskip 
\noindent
The main contributions of this work are as follows:
\begin{itemize}
    \item A reasoning-centric multi-agent routing system. We introduce TCAR, the first domain router that outputs natural-language decision rationales and supports flexible multi-agent selection.
    \item Conflict resolution through collaborative execution: We propose an innovative "multi-agent generation and refinement" pipeline that mitigates agent conflicts and effectively integrates complementary domain expertise.
    \item An open-source collaborative routing framework. We release the model and associated resources to advance research in interpretable multi-agent routing and reasoning-driven LLM collaboration.
\end{itemize}

\section{Related Work}

Performance based routing selects among models of different sizes to balance cost and latency. Hybrid LLM\cite{ding2024hybrid} proposes a model based on the BERT architecture to assess the difficulty of a query and make routing decisions. Self-REF\cite{chuang2024learning} uses an LLM to output a confidence score, For low confidence, it uses an LLM or rejects the answer, while for high confidence, it uses an SLM. 

\medskip 
\noindent
Task based routing assigns queries to domain-specific expert agents to achieve higher accuracy. This work focuses on task based routing, whose methodological landscape can be grouped into several major directions. Early approaches largely employed BERT-style encoders \cite{simonds2024modem}, formulating routing as a fixed single-label classification problem grounded in a predefined intent space and model structure. To support adding new agents without retraining, some studies shifted toward KNN-based methods, such as computing similarity between query and agent representations\cite{piskala2025dynamic}, or using query–agent dual-tower models or fusion-based cross-encoders \cite{chiu2022cross} to dynamically compute relevance scores. \cite{shnitzer2023large}Train a binary classifier to determine whether the model performs excellently on a specific task. More recently, LLM-based routing methods have emerged—for example, ArchRouter \cite{tran2025arch} leverages human preference to perform agent selection. AgentRouter\cite{zhang2025agentrouter} Transform MAS into a knowledge graph, and decide which agent to select by predicting the edges through the model. Confidence-Driven LLM Router\cite{zhang2025leveraging} decide which agent is better based on the confidence of the responses output by different agents, thereby obtaining labeled preference data, which is then used to train the routing model. However, these methods still follow the paradigm of producing a single agent output, lacking explicit and structured reasoning explanations. As a result, they struggle to handle the complexity and interpretability demands in real enterprise environments where multiple agents may simultaneously be relevant.

\medskip 
\noindent
Multi-agent collaboration frameworks have also advanced rapidly in recent years. Representative systems include AutoGen \cite{wu2024autogen}, CAMEL \cite{li2023camel}, and CrewAI \cite{CrewAI}, which enable multiple agents to jointly solve complex tasks through mechanisms such as message passing, role-playing, and multi-turn dialogues. However, these frameworks typically rely on manually specified participant agents or simple rule-based selection. Their focus lies in how agents collaborate to execute a task rather than which agents should participate. Consequently, they lack a systematic mechanism for identifying the relevant subset of expert agents from a larger pool, and they pay limited attention to how the outputs of multiple agents should be integrated into a coherent final answer.

\section{Method}

\noindent
We present TCAR, an adaptive reasoning router for multi-agent collaboration

\subsection{Problem Formulation}

\noindent
A multi-agent system consists of a set of expert agents with different domain specializations. Let $A=\{a_1,a_2,...,a_N\}$ denote the set of all available expert agents. In real systems, each expert agent is not hard-coded into the router through a fixed identifier; instead, it exposes its capabilities and responsibilities to the routing model through a natural-language description. We denote this description as $d_i$, and define a natural-language "encoding function" $g$ such that $g(a_i)=d_i$.The set of all descriptions is $D=\{d_1,d_2,...,d_N\}$ Given a user query $q$, the router does not operate directly on $q$ alone. Instead, it receives a full prompt formed by concatenating the routing instruction, the user query, and all agent descriptions. Formally, we define a prompt construction function:
$$Prompt(q,A)=\Phi(ins,q,D)$$
where $\Phi$ denotes ordered text concatenation and $ins$ is the router instruction.
The routing task is then to output both the reasoning $C_q$ for agent selection and the final selected agent set $A_q$:
$$\mathcal{R}: Prompt(q,A) \rightarrow (C_q, A_q)$$

\subsection{Dynamic Agent Addition}

\noindent
Traditional methods typically bind a router $\mathcal{R}$ to a fixed agent set $A$, when $A$ changes $\mathcal{R}$ must be retrained. In TCAR, adding a new agent requires only appending $a_{new}$ to the existing set, yielding $A'=A \cup \{a_{new}\}$ which produces a new description set $D'=D \cup \{d_{new}\}$, The routing process then becomes $(C_q,A_q)=\mathcal{R}(Prompt(q,D')),A_q \subseteq A$, meaning that the router $\mathcal{R}$ itself does not need to be modified or retrained to support newly added agents. This property greatly improves efficiency and usability in real-world enterprise environments.

\subsection{Multi-Agent Conflicts}

\noindent\textbf{Selecting Multiple Agents.} Traditional task routing typically formulates routing as a single-label classification problem:$\mathcal{R}(Prompt(q,A))=a_i \in A$ meaning that each query can only be assigned to a single agent. This formulation implicitly assumes strict and mutually exclusive domain boundaries, as well as a clearly defined intent space—assumptions that rarely hold in real enterprise environments. When multiple agents are simultaneously suitable for handling a query due to overlapping domain responsibilities or cross-domain problem characteristics, an agent conflict occurs, formally defined as: $|A_q|>1$. Existing single-label routers cannot represent this structural property; they are forced to pick only one agent among multiple plausible candidates. This leads to higher routing error rates, brittle behavior under cross-domain or ambiguous queries, and a lack of interpretability in routing decisions—ultimately undermining system stability and trustworthiness. To address this, we redefine routing as selecting a subset of agents from the full agent set: $A_q \subseteq A, |A_q|\ge1$. The objective is no longer to choose a single "correct" agent, but to identify all experts that may be relevant to the query.

\medskip  
\noindent\textbf{Reasoning-Based Agent Selection.} To effectively model multi-agent conflict scenarios, we reformulate routing from a “direct label prediction” task into a two-stage “reason–then–select” process. Specifically, instead of outputting only a discrete agent label, the router generates both a natural-language reasoning chain $C_q$ and the corresponding agent subset $A_q$:
$$(C_q, A_q)=\mathcal{R}(Prompt(q,A))$$
The reasoning $C_q$ is required to explicitly analyze the potential causes of the user query, the relevant technical stack, and the responsibility boundaries of each agent. This produces a textual explanation of why the selected agents are relevant to the current query. This design provides two key advantages. First, the reasoning process encourages fine-grained semantic alignment between the query and agent descriptions, leading to more stable selection of the appropriate agent subset in cases of overlapping responsibilities or semantic ambiguity—rather than relying on brittle short-text matching. Second, explicit reasoning enhances interpretability: operations engineers can inspect $C_q$ to diagnose routing errors, refine agent descriptions, or adjust routing strategies, forming a closed feedback loop of "data → reasoning → policy". Experimental results show that, compared with black-box single-label or multi-label routers, incorporating natural-language reasoning significantly improves routing robustness under cross-domain, weakly-specified, and noisy queries.

\medskip  
\noindent
\textbf{Multi-Agent Collaborative Execution.} Once the router is allowed to output multiple agents, potential inter-agent conflicts no longer need to be "compressed" into a single decision during routing. Instead, they are explicitly preserved and leveraged downstream through collaborative execution. Given the routing result $A_q=\{a_{i_1},\dots,a_{i_k}\},\ k=|A_q|$,the system invokes all selected candidate agents in parallel, with each agent producing a response to the same query $r_{i_j}=a_{i_j}(q)$. On top of these responses, we design a subsequent collaboration and aggregation mechanism. On the one hand, different agents provide partial yet specialized answers from their own perspectives, helping to cover diverse potential root causes in cross-domain queries. On the other hand, a downstream aggregation module—the Refining Agent—filters, aligns, and fuses$\{r_{i_j}\}$into a single consistent final response. Formally, this collaborative process is defined as:
$$y_q = \mathcal{F}(\{r_{i_j} \}_{j=1}^k)$$
where $\mathcal{F}$ denotes the overall operator of multi-agent collaboration and aggregation. Through the design of multi-agent selection + collaborative execution + aggregated fusion, the system no longer relies on a one-shot single-point decision to eliminate conflicts. Instead, conflicts are explicitly transformed into multi-perspective evidence that is unified at the answer level, thereby improving final response accuracy and robustness while maintaining high coverage.

\begin{figure}[t]
    \centering
    \includegraphics[width=1\linewidth]{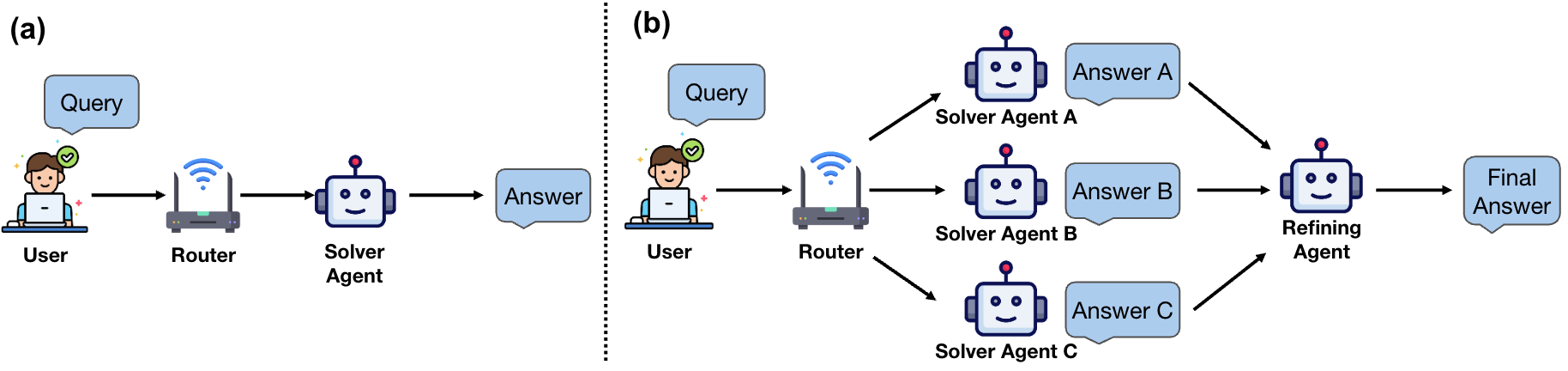}
    \small\caption{(a) Solo Agent: the router outputs a single agent, and that agent directly responds to the user's query.(b) Refining Agent: the router outputs a set of candidate agents, and the refining agent integrates their outputs into a final answer.}
    \label{fig:agent}
\end{figure}

\subsection{Training TCAR}
To endow the router $\mathcal{R}$ with both multi-agent selection capabilities and interpretable reasoning, we adopt a two-stage\cite{ouyang2022training} training strategy. We first apply supervised fine-tuning (SFT)\cite{chung2024scaling} to teach the model the basic “reason–then–select” pattern. We then further enhance the quality of agent selection and the stability of reasoning through reward-based reinforcement learning using DAPO \cite{yu2025dapo}.

\medskip  
\noindent\textbf{Supervised Fine-Tuning.} In the first stage, we train the model using labeled data with standard causal language modeling. SFT equips the model with the following capabilities: (i) performing semantic alignment between the query and agent descriptions, thereby improving its generalization ability when dynamically onboarding new agents; and (ii) generating structured outputs, including the reasoning chain $C_q$ and the candidate agent set $A_q$. Instead of adopting the conventional <think> tag used in “think”-style models, we introduce a dedicated <reason> tag to ensure compatibility with both instruction-following models and think models. The SFT stage is optimized using the standard autoregressive language modeling loss:
$$
\mathcal{L}_{\mathrm{SFT}}(\theta)
= -\sum_{t=1}^{T} \log p_\theta(y_t \mid y_{<t}, q, A),
$$
$$
\{y_1,\dots,y_T\} = (C_q, A_q).
$$

\noindent However, SFT also has its limitations: the model may overfit to annotation templates or generate reasoning in a mechanical manner, making it necessary to further improve both the quality of reasoning and the accuracy of agent selection.

\medskip
\noindent\textbf{Reinforcement Learning.}
To improve the robustness of routing decisions and the soundness of the generated reasoning, we introduce a reinforcement learning stage on top of SFT. Here, we adopt the DAPO method to train the model, with the training objective formulated as:
\[
\begin{aligned}
& \mathcal{J}_{\mathrm{DAPO}}(\theta)= \mathbb{E}_{(q, a) \sim \mathcal{D},\left\{o_i\right\}_{i=1}^G \sim \pi_{\theta_{\text {old }}}(\cdot \mid q)} \\
& {\left[\frac{1}{\sum_{i=1}^G\left|o_i\right|} \sum_{i=1}^G \sum_{t=1}^{\left|o_i\right|} \min \left(r_{i, t}(\theta) \hat{A}_{i, t}, \operatorname{clip}\left(r_{i, t}(\theta), 1-\varepsilon_{\text {low }}, 1+\varepsilon_{\text {high }}\right) \hat{A}_{i, t}\right)\right] } \\
& \text{ s.t. } 0< | \{o_i | \text{ is\_equivalent }(a, o_i)\} | <G .
\end{aligned}
\]

\noindent For reward design\cite{deepseekai2025deepseekr1incentivizingreasoningcapability}, since the SFT stage already enables the model to produce well-formatted outputs with a reasonable reasoning length $C_q$, we apply the reward only to the final agent set $A_q$. Because $A_q$ is a set that may contain one or multiple agents, we decompose the routing reward into two parts. Let $A^*$ denote the gold agent set and $A_q$ the predicted candidate agent set. The first reward focuses on whether each element in $A_q$ belongs to $A^*$ (analogous to precision), while the second reward evaluates whether $A_q$ covers all correct agents (analogous to recall or a coverage constraint). Formally, the rewards are defined as follows:

\medskip 
\noindent
First, we define a local reward based on the degree of set overlap, which measures the proportion of agents in the predicted set $A_q$ that are correct. This reward encourages the model to reduce irrelevant or incorrect agent predictions.
\[
R_1(A_q, A^*) = \frac{|A_q \cap A^*|}{\max(1, |A_q|)},
\]

\medskip 
\noindent
Second, we introduce a coverage reward to penalize missing correct agents. The model receives an additional reward if and only if its predicted set $A_q$ fully contains the ground-truth set $A^*$; otherwise, this reward is zero.
\[
R_2(A_q, A^*) = \frac{|A_q \cap A^*|}{\max(1, |A^*|)},
\]

\medskip 
\noindent
To prevent the model from generating excessively long outputs or endlessly listing agents, we introduce an additional length-penalty term\cite{tan2025gtpo}. Combining all three components, we define the final reward as:
\[
R(A_q, A^*) = \alpha \, R_1(A_q, A^*) + (1-\alpha) R_2(A_q, A^*) - \beta max(|A_q| - |A^*|, 0)
\]

\medskip 
\noindent
where $\alpha$ and $\beta$ are weighting hyperparameters. The coefficient $\alpha$ balances the trade-off between the correctness of the predicted set (precision-like) and its coverage of the ground-truth set (recall-like), while $\beta$ controls the strength of the length-penalty term. In practice, we directly incorporate this reward into the policy optimization objective during the reinforcement learning stage, encouraging the router to avoid over-predicting irrelevant agents while still covering all truly relevant experts in multi-agent scenarios.

\section{Experiments}

We present the experimental results and analyses.

\medskip
\subsection{Experimental Setup}
This section introduces the datasets, baselines, evaluation metrics, and implementation details used to evaluate TCAR. All experiments focus on the core task of multi-agent domain routing, aiming to thoroughly assess TCAR’s effectiveness under domain conflicts, ambiguous intents, and real-world enterprise scenarios.

\medskip 
\noindent
\textbf{Datasets.} We conduct evaluations on four public intent-classification datasets and one internal enterprise dataset:
\begin{itemize}
\item \textbf{CLINC150} \cite{larson2019evaluation}: Contains 150 intent categories covering everyday conversational scenarios. The task boundaries are clear, but the large number of classes leads to long agent descriptions, posing challenges for LLMs.
\item \textbf{HWU64} \cite{liu2021benchmarking}: Includes 64 intents with relatively high cross-domain ambiguity.
\item \textbf{MINDS14} \cite{DBLP:journals/corr/abs-2104-08524}: A multilingual dataset spanning 14 domains, used to test router stability under cross-lingual semantic transfer.
\item \textbf{SGD} \cite{rastogi2020schema}: A multi-turn dialogue dataset designed to evaluate routing performance in multi-turn conversational settings.
\item \textbf{QCloud}: Additionally, we construct a real-world enterprise dataset from Tencent Cloud, covering domains such as networking, cloud storage, CDN, security, databases, and application operations. Compared with public datasets, it exhibits stronger domain coupling, higher intent ambiguity, and more frequent agent conflicts, making it a crucial benchmark for assessing the practicality of multi-agent routing systems.
\end{itemize}

\medskip 
\noindent
\textbf{Evaluation Metrics.} To comprehensively evaluate routing performance, we adopt the following metrics:
\begin{itemize}
\item For single-agent datasets, we primarily report \textbf{accuracy}. For multi-agent datasets, we compute \textbf{F1} to measure whether the model successfully and accurately identifies all agents capable of handling a given query.
\item \textbf{End-to-End Task Success Rate}, which evaluates the overall response quality after multi-agent collaboration and Refining Agent processing, offering a system-level assessment of routing capability.
\end{itemize}

\noindent 
\textbf{Baselines.} Since static label routing is difficult to deploy in most real-world enterprise scenarios, we focus on comparing against dynamic routing approaches. We evaluate a variety of both proprietary and open-source large language models, including GPT-5.1\cite{openai2025gpt51}, Claude-4.5\cite{anthropic2025sonnet45}, DeepSeek-v3.1\cite{deepseekai2024deepseekv3technicalreport}, ArcRouter\cite{tran2025arch}, as well as the Qwen3\cite{qwen3technicalreport} family, which also serves as our base model for training.

\medskip
\noindent 
\textbf{Implementation Details.} We experimented with several model sizes from the Qwen3 family and ultimately selected Qwen3-4B-Instruct-2507 for open-sourcing, achieving a strong balance between effectiveness and efficiency. To ensure training data compatibility and avoid reliance on model-specific reasoning formats, we did not use Qwen3’s native \texttt{<think>} tag. Instead, we introduced a unified \texttt{<reason>} tag for generating reasoning chains. Full-parameter SFT was conducted using the ms-swift framework\cite{zhao2024swiftascalablelightweightinfrastructure} with data parallelism across 8 GPUs, Adam optimizer, batch size of 256, training for 1 epoch, an initial learning rate of 2e-5, and a warmup ratio of 0.1. During the RL stage, we use the SFT model with a temperature of 1 and perform 8 rollouts per training sample. If the rollout consistency exceeds $\tau > 0.6$, we consider the sample to be already well-learned during SFT—essentially a low-entropy token \cite{wang2025beyond}. Such samples tend to cause near-zero variance when computing advantages during RL, leading to ineffective optimization. Therefore, following the principles of \cite{yu2025dapo}, we remove these samples from the RL training set to reduce unnecessary dynamic sampling and improve training efficiency. We also experimented with initializing the model using Slerp\cite{shoemake1985animating,mlabonne_merge_models_2025}, which allows the model to achieve higher pass@k performance at the early stage of RL training. We aim for the model in the RL phase to transition from achieving high pass@k to achieving high pass@1\cite{walder2025pass}. Thereby preventing the model from suffering entropy collapse\cite{cui2025entropy}. Both stages of training (SFT and RL) are implemented using the ms-swift framework.

\subsection{Main Results}

\medskip
\noindent
Table~\ref{tab:router-results} presents the performance of all models across the five datasets (CLINC150, HWU64, MINDS14, SGD, QCloud). Overall, general-purpose LLMs such as GPT-5.1 and DeepSeek-v3.1 outperform most small-scale open-source models by a considerable margin. Despite having only 4B parameters, our TCAR achieves state-of-the-art performance on all datasets except CLINC150. CLINC150 contains 150 intent categories, resulting in extremely long prompts (averaging 18k tokens), which pose challenges for smaller models that struggle with ultra-long sequence processing. On multi-turn (SGD) and multilingual (MINDS14) datasets, TCAR demonstrates clear and substantial advantages. On the QCloud dataset which features high agent conflict frequency and real enterprise ambiguity—TCAR surpasses even current leading general-purpose LLMs in F1, highlighting its robustness in multi-agent routing scenarios.

\begin{table}[ht]
\centering
\begin{footnotesize}
\begin{tabular}{lccccc}
\toprule
\textbf{Models} & \textbf{CLINC150} & \textbf{HWU64} & \textbf{MINDS14} & \textbf{SGD} & \textbf{QCloud}\\
\midrule
GPT-5.1                 & 93.84 & 85.59 & 95.59 & 73.90 & 92.93\\
Claude-Sonnet-4.5       & \textbf{94.21} & 87.40 & 96.20 & 76.02 & 91.45\\
DeepSeek-v3.1-terminus  & 88.29 & 88.10 & 95.72 & 79.70 & 92.98\\
ArcRouter               & 62.98 & 69.33 & 91.79 & 65.59 & - \\
Qwen3-Embedding-4B      & 57.21 & 54.27 & 94.12 & 37.02 & - \\
Qwen3-4B-Instruct-2507  & 70.12 & 80.29 & 90.08 & 58.74 & 80.81\\
\midrule
\textbf{TCAR(4B)}       & 91.25 & \textbf{91.63} & \textbf{96.70} & \textbf{91.58} & \textbf{93.98}\\
\bottomrule
\end{tabular}
\end{footnotesize}
\small\caption{Comparison of model performance across datasets. For the QCloud dataset, the metric is reported as F1, while for all other datasets the metric is Accuracy.}
\label{tab:router-results}
\end{table}

\begin{figure}[t]
    \centering
    \includegraphics[width=0.7\linewidth]{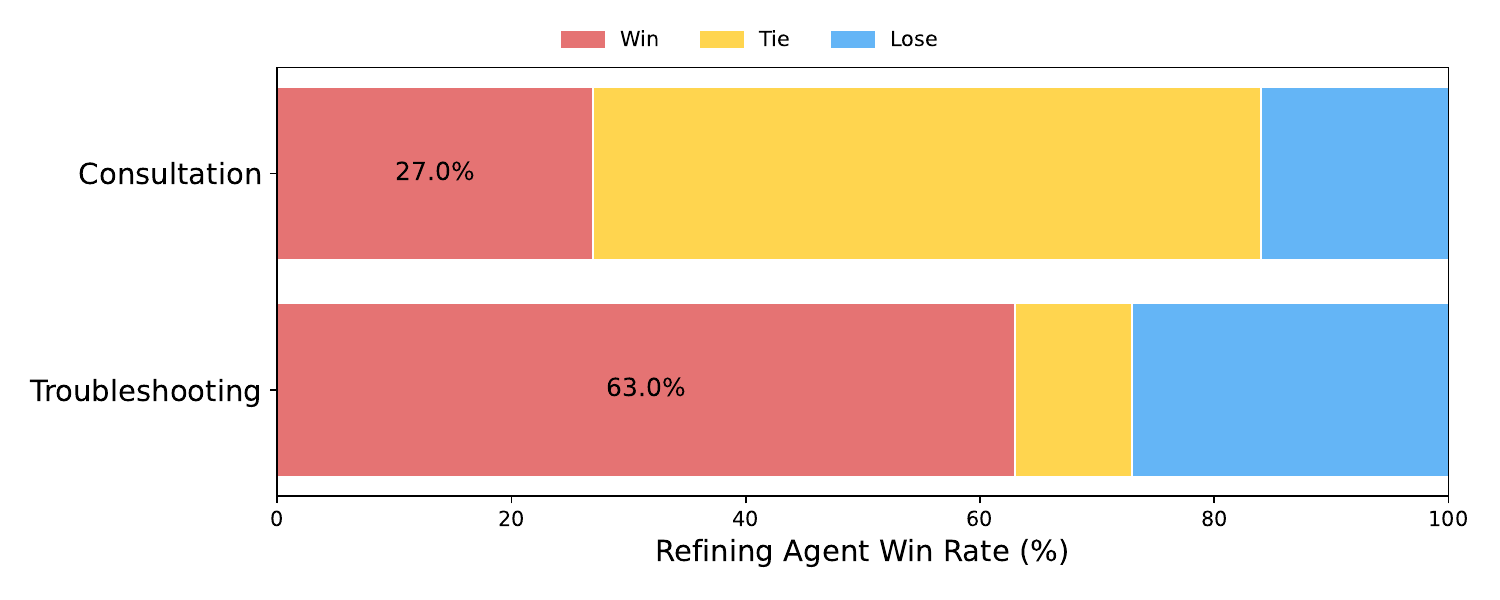}
    \small\caption{In scenarios where consultation-type and troubleshooting-type queries create conflicts among candidate agents, we compare two strategies: randomly selecting a single agent to respond versus allowing multiple agents to answer followed by aggregation by the Refining Agent. The results show that the Refining Agent achieves a significantly higher human preference win rate, particularly for troubleshooting-type queries.}
    \label{fig:win-rate}
\end{figure}

\medskip
\noindent
We further evaluate the effectiveness of the downstream Refining Agent when TCAR outputs multiple candidate agents. As shown in Figure~\ref{fig:win-rate}, the Refining Agent implemented using DeepSeek-v3.1 as the underlying LLM—significantly enhances the quality of response in our QCloud data set. For consultation-type queries, a single agent is often sufficient to produce high quality responses. However, for troubleshooting-type queries, an individual agent typically fails to provide complete coverage, whereas the Refining Agent integrates and consolidates responses from multiple agents, resulting in more comprehensive and reliable answers.

\subsection{Ablation Studies}
To systematically analyze the contribution of each component in TCAR, we conduct ablation studies along three dimensions: reasoning capability, training strategy, and the source of RL initialization.

\medskip
\noindent 
\textbf{Effect of Reasoning}
This experiment evaluates the impact of explicit reasoning chains\cite{NEURIPS2022_9d560961} on routing performance.

\begin{itemize}
\item \textbf{Without reasoning}: The model performs direct classification prediction and outputs only $A_q$.
\item \textbf{With reasoning}: The model first generates $C_q$ as an explicit and interpretable reasoning chain and then outputs the set of agents $A_q$.
\end{itemize}

\medskip
\noindent
We evaluated both settings on the public datasets and the QCloud dataset. The results~\ref{tab:think} show that even without reasoning, the model achieves strong performance after fine-tuning; however, incorporating reasoning consistently leads to further improvements. This observation suggests that reasoning enhances the model’s generalization ability.

\begin{table}[ht]
\centering
\begin{footnotesize}
\begin{tabular}{lccccc}
\toprule
\textbf{Reasoning Setting} &
\textbf{CLINC150} & \textbf{HWU64} & \textbf{MINDS14} &
\textbf{SGD} & \textbf{QCloud} \\
\midrule
TCAR (without Reasoning) & 89.65 & 87.83 & 96.45 & 84.96 & 91.33 \\
TCAR (with Reasoning)    & \textbf{91.25} & \textbf{91.63} & \textbf{96.70} & \textbf{91.58} & \textbf{93.98} \\
\bottomrule
\end{tabular}
\end{footnotesize}
\small\caption{Comparison of the performance with and without reasoning output}
\label{tab:think}
\end{table}

\medskip
\noindent 
\textbf{SFT-Only vs. SFT+RL}
This experiment evaluates the improvements brought by applying DAPO during the RL stage.

\begin{itemize}
\item \textbf{SFT-only}: The model is trained using supervised fine-tuning only.
\item \textbf{SFT+RL}: After SFT, we further filter the training set and retain only high-entropy tokens for RL.
\end{itemize}

\noindent 
We compare the performance of the SFT-only model with the model further optimized through RL. Experimental results show that RL leads to consistent and significant improvements. In particular, on the QCloud dataset, RL enhances recall while maintaining high precision, effectively mitigating the common issue of SFT-only models being overly conservative and outputting only a single agent.

\begin{table}[ht]
\centering
\begin{small}
\begin{tabular}{lccccc}
\toprule
\textbf{Training Method} &
\textbf{CLINC150} & \textbf{HWU64} & \textbf{MINDS14} &
\textbf{SGD} & \textbf{QCloud} \\
\midrule
SFT-Only & \textbf{91.32} & 90.30 & 96.69 & 86.30 & 93.77 \\
SFT+RL   & 91.25 & \textbf{91.63} & \textbf{96.70} & \textbf{91.58} & \textbf{93.98} \\
\bottomrule
\end{tabular}
\end{small}
\small\caption{Comparison between SFT-only training and SFT followed by RL.}
\label{tab:sft-rl}
\end{table}

\medskip
\noindent 
\textbf{RL Initialization: SFT-Only vs. SFT+Slerp}
Since the initialization point of RL has a substantial impact on the final performance, we compare two different SFT models used for RL initialization:

\begin{itemize}
\item \textbf{SFT-Only}: RL initialized from the SFT model.
\item \textbf{SFT+Slerp)}: RL initialized from the model obtained via SFT-Slerp merging.
\end{itemize}

\begin{figure*}[t]
    \centering
    \begin{subfigure}{0.49\linewidth}
        \centering
        \includegraphics[width=\linewidth]{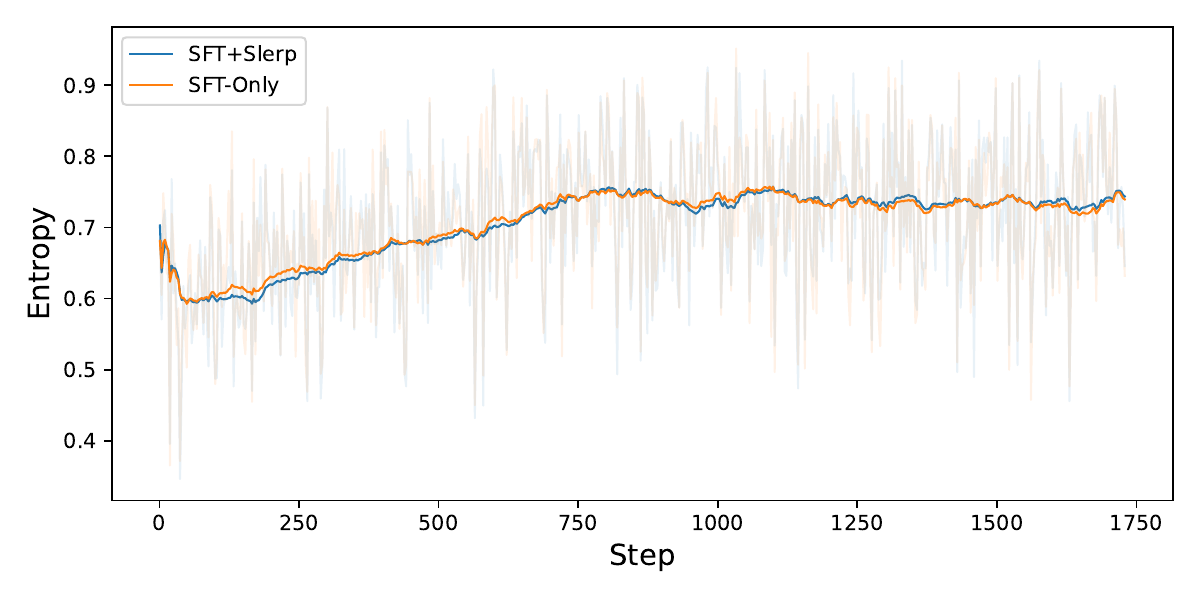}
        \caption{RL Reward}
        \label{fig:reward}
    \end{subfigure}
    \hfill
    \begin{subfigure}{0.49\linewidth}
        \centering
        \includegraphics[width=\linewidth]{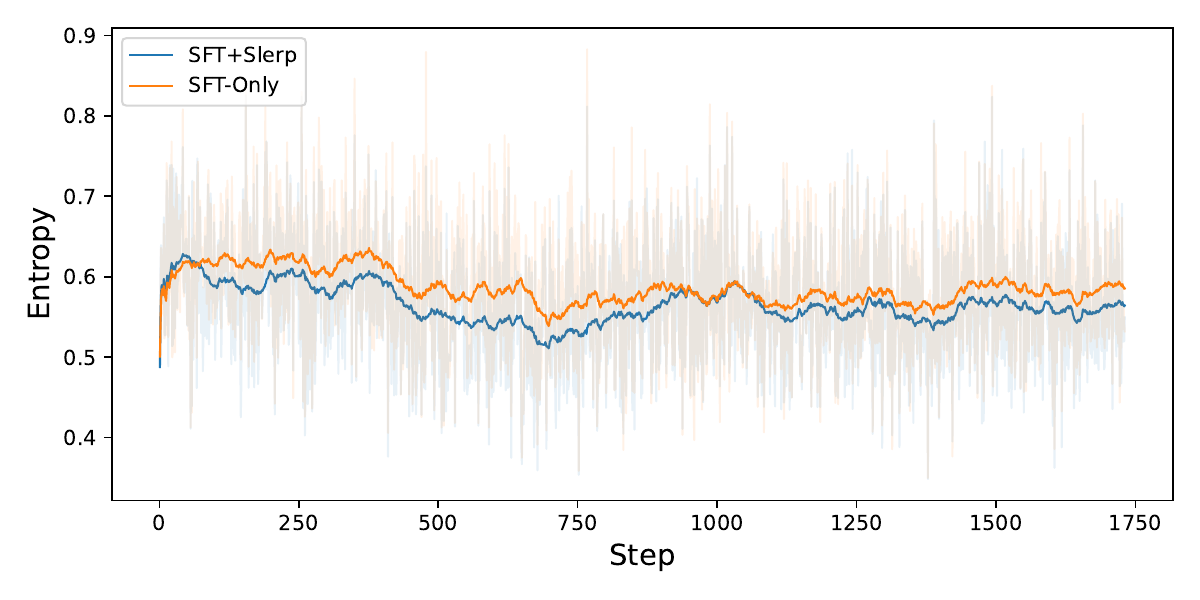}
        \caption{RL Entropy}
        \label{fig:slerp}
    \end{subfigure}
    \small\caption{Comparing RL initialized from the SFT model and from the Slerp-merged model, we observe that both approaches achieve almost the same reward throughout training. In contrast, the Slerp-initialized model maintains higher entropy, suggesting that it explores a broader solution space.}
    \label{fig:two_curves}
\end{figure*}

\noindent 
Slerp is a model-merging technique \cite{shoemake1985animating,mlabonne_merge_models_2025} that effectively enhances model generalization. We trained five different SFT models on five distinct datasets and then merged them using Slerp. The resulting merged model exhibits higher exploration capability in the early stages of RL, reflected by a higher pass@k score. We aim for the model in the RL phase to transition from achieving high pass@k to achieving high pass@1\cite{walder2025pass}. We conducted RL training using both the SFT model and the Slerp-merged model. As shown in Table~\ref{tab:slerp}, the Slerp-based model does not exhibit a significant improvement in the final metrics. However, from Figure~\ref{fig:slerp}, we observe that although the reward curves of the two initialization strategies are similar, the Slerp model maintains a higher entropy, indicating stronger exploration. This enhanced exploration leads to better generalization and provides greater potential for further fine-tuning based on the merged model.

\begin{table}[ht]
\centering
\begin{small}
\begin{tabular}{lccccc}
\toprule
\textbf{RL Initialization} &
\textbf{CLINC150} & \textbf{HWU64} & \textbf{MINDS14} &
\textbf{SGD} & \textbf{QCloud} \\
\midrule
SFT-Only  & 91.25 & 91.63 & 96.70 & 91.58 & 93.97 \\
SFT+Slerp     & 90.43 & 91.45 & 96.82 & 91.53 & 93.20 \\
\bottomrule
\end{tabular}
\end{small}
\small\caption{Comparison between initializing the RL stage with the SFT model and initializing it with the Slerp-merged model.}
\label{tab:slerp}
\end{table}

\medskip
\noindent \textbf{Summary of Ablation Findings}

\noindent Based on the ablation results, we summarize the following conclusions:
\begin{itemize}
\item Reasoning capability is a key factor for improving routing interpretability and robustness.
\item RL is an essential training stage that enhances recall while maintaining high precision.
\item The choice of RL initialization point is crucial; Slerp-based models offer stronger generalization and are less prone to entropy collapse during RL.
\end{itemize}

\section{Analysis}
\noindent After validating the overall effectiveness of TCAR through the main results and ablation studies, we further analyze its behavioral characteristics and internal mechanisms from multiple perspectives. This section provides an in-depth examination of error types, reasoning quality, multi-agent collaboration effects, and system efficiency, offering a more comprehensive understanding of the model's strengths and limitations.

\subsection{Error Analysis}
To systematically understand the major sources of routing errors, we categorize the mistakes and observe that TCAR's errors primarily fall into three types: (1) \textbf{Incomplete problem descriptions}. User queries may lack key information. For example, in “What should I do if my webpage loads slowly?”, although TCAR’s reasoning often infers that multiple agents are needed, the lack of essential context may still lead to deviations; see Appendix~\ref{badcase:1}. (2) \textbf{Insufficient understanding of highly domain-specific scenarios}. In Tencent Cloud use cases, the model sometimes struggles with understanding specialized terminology, leading to incorrect routing decisions; see Appendix~\ref{badcase:2}. (3) \textbf{Mismatch between reasoning and final prediction}. Some cases exhibit reasonable reasoning steps but produce agent outputs inconsistent with the reasoning itself. How to further improve the alignment and consistency of the reasoning chain remains an open problem.

\subsection{Multi-Agent Collaboration Analysis}
Multi-agent collaboration effectively supplements information from different perspectives. In Tencent Cloud troubleshooting scenarios, multiple agents are often required to jointly diagnose and resolve issues, as a single agent is typically insufficient. Moreover, the Refining Agent demonstrates strong capability in reconciling conflicting responses. With RAG support, the Refining Agent compares and filters multiple candidate answers using the knowledge base, enabling conflicting opinions to converge. For instance, in troubleshooting scenarios localization tasks, partial correctness from different agents can be combined to form a complete solution—something a single agent cannot achieve.

\subsection{Efficiency and Cost Analysis}
We further evaluate the deployment cost of TCAR in online environments. (1) \textbf{The number of selected agents is far below the theoretical maximum}. Although multi-agent outputs are supported, the actual average number of routed agents is only 1.37, indicating that most queries can be solved by a single agent and that the model naturally favors concise predictions. (2) \textbf{The latency impact of the reasoning chain is well controlled}. For a 4B model deployment requirements remain modest; the average reasoning length is under 100 tokens. When deployed on GPUs, the average latency is less than 1 second, which is entirely acceptable. (3) \textbf{The incremental cost of downstream collaboration is small}. Because the number of selected agents is low, downstream execution incurs an average overhead equivalent to only 0.37 additional agent calls—far below any risk of combinatorial explosion.

\section{Limitations}
Although TCAR demonstrates stable and significant performance improvements across multiple public datasets and real-world enterprise scenarios, several limitations remain that warrant attention in future work.

\medskip
\noindent
\textbf{Dependence on the quality of agent descriptions.}  
TCAR's routing reasoning relies on manually written agent descriptions, which serve as the semantic foundation for constructing reasoning chains. When these descriptions are overly brief, ambiguous, or fail to fully represent the true scope of an agent’s responsibilities, the resulting reasoning chain may deviate from the intended semantics, ultimately affecting routing accuracy. Although reinforcement learning improves the model’s robustness to noise in descriptions, the overall quality of these descriptions remains a critical factor influencing performance.

\medskip
\noindent
\textbf{Challenges in long-tail knowledge and domain transfer.}  
For low-frequency or highly domain-specific scenarios (e.g., rare APIs or special network configurations), the model may still exhibit instability due to insufficient training samples. Furthermore, when the instruction format is heavily modified, the model’s instruction-following capability may degrade. If the instructions are substantially altered, an additional round of fine-tuning on domain-specific private data becomes a more reasonable strategy.

\medskip
\noindent
Overall, these limitations do not undermine TCAR’s effectiveness in mainstream routing tasks or real enterprise deployments, but they highlight promising directions for future research aimed at improving robustness, generalization, and performance under extreme or specialized conditions.

\section{Conclusion}
This paper addresses the prevalent challenges in real-world multi-agent enterprise systems, including agent conflicts, cross-domain intent ambiguity, and limited interpretability. We propose \textbf{TCAR}, a reasoning-centric multi-agent routing framework. By explicitly generating a natural-language reasoning chain, TCAR extends traditional single-agent classification into a flexible subset prediction over all potentially relevant agents, significantly enhancing the coverage and interpretability of routing decisions. Building on this, we further design a downstream multi-agent collaborative execution mechanism and a Refining Agent that consolidates multiple agent outputs, enabling the system to harness complementary expertise across complex or cross-domain tasks.Experiments on multiple public datasets and large-scale real ITSM data from Tencent Cloud demonstrate that TCAR achieves substantial improvements in task accuracy, recall under conflict-prone scenarios, and the final answer quality resulting from multi-agent collaboration. Ablation studies further validate the effectiveness of reasoning chains, reinforcement learning, and the SLERP-based model fusion strategy. Meanwhile, our analyses reveal key factors influencing model performance, including reasoning-chain structure, agent description quality, and the sparsity of long-tail domain knowledge.In summary, TCAR highlights the feasibility and practical value of a reasoning-driven, multi-agent collaborative routing paradigm in large-scale enterprise environments. For future work, we plan to explore structured reasoning-chain constraints, more efficient collaboration protocols, and low-cost expansion strategies for emerging business domains to further enhance robustness and generalization in complex real-world scenarios.

\clearpage
\bibliographystyle{unsrt}
\bibliography{refs}

\clearpage
\appendix
\section{prompt}

\textbf{TCAR Prompt:}
\begin{tcolorbox}
\small
You are a Router Agent. Your job is to determine which Agents should handle the user's issue based on the chat history.Your goal is to select the appropriate Agents to ensure the problem can be solved efficiently and accurately. You may select up to 3 Agents.

\# Agent List

\{agents\}
\medskip

\# Notes

+ The Description is each Agent’s capability description. When selecting an Agent, you must ensure that its capabilities clearly allow it to resolve the user’s problem.

+ If no Agent can reasonably handle the problem, output "oos" in the <ID> tag.
\medskip

\# Output Format

Output in XML format. If multiple Agents are needed, output multiple <ID> lines.

<Reason>The reason for choosing the Agent(s)</Reason>

<ID>Agent ID</ID>
\medskip

\# Chat History
\end{tcolorbox}

\noindent\textbf{Refining Agent Prompt:}
\begin{tcolorbox}
\small
You are a Refining Agent. 
I will provide you with a user query along with the answers generated by different agents. Your task is to compare and integrate these answers and produce the most helpful and accurate final response for the user.

\# Requirements

Please output one final answer that satisfies the following criteria:

+ Fully compare the content from different agents, and combine the strengths of each answer to produce the response that best solves the user’s problem. Avoid repetition, errors, and irrelevant information.

+ If there are conflicts among the answers, you must present all conflicting points in the final response and explicitly explain the conflict.

+ If the answers are complementary, you should integrate them into a unified, coherent response.

+ If all answers fail to provide useful information or all refuse to answer the query, you must output a polite refusal to answer.

+ Output only the final answer — do not include any other commentary.
\medskip

\# Query: 
\{query\}

\# Answers:
\{answers\}
\medskip

\# Final Answer:
\end{tcolorbox}

\section{Badcase}

\textbf{Incomplete problem descriptions}
\begin{tcolorbox}
\label{badcase:1}
\small
Input Query:
What should I do if a webpage loads very slowly?

TCAR Output:

<Reason>
The user reports that “the webpage loads very slowly,” which requires diagnosis from both the network path and acceleration layers. The Network Agent can analyze public/private network latency, packet loss, jitter, and path quality. The CDN Agent can examine CDN access latency, origin fetch behavior, and distribution policies. If EdgeOne is being used, the EO Agent should investigate origin latency, edge node issues, and configuration problems.
</Reason>

<ID>network\_agent</ID>

<ID>cdn\_agent</ID>

<ID>eo\_agent</ID>
\end{tcolorbox}

\noindent\textbf{Insufficient understanding of highly domain-specific scenarios}
\begin{tcolorbox}
\label{badcase:2}
Agent List:
\begin{lstlisting}
<Agent>
    <Name>CKafka Message Queue Agent</Name>
    <Description>
        Query CKafka instance information and diagnose CKafka-related issues.</Description>
    <ID>ckafka_agent</ID>
</Agent>

<Agent>
    <Name>RocketMQ Message Queue Agent</Name>
    <Description>
        Analyze abnormal RocketMQ topic production rates and query configuration changes (scaling/reconfiguration).
    </Description>
    <ID>rocketMQ_agent</ID>
</Agent>
\end{lstlisting}
Input Query:
Configuration change for CKafka message queue

TCAR Output:

<Reason>
The user reports a “message queue configuration change.” The RocketMQ Message Queue Agent has the capability to handle configuration change queries and can effectively address the user’s issue.
</Reason>

<ID>rocketMQ\_agent</ID>

\end{tcolorbox}

\end{document}